# Towards Best Practice of Interpreting Deep Learning Models for EEG-based Brain Computer Interfaces

Jian Cui[1], Liqiang Yuan[2], Zhaoxiang Wang[1*], Ruilin Li[2*], Tianzi Jiang[1,3]

***Abstract*—As deep learning has achieved state-of-the-art performance for many tasks of EEG-based BCI, many efforts have been made in recent years trying to understand what have been learned by the models. This is commonly done by generating a heatmap indicating to which extent each pixel of the input contributes to the final classification for a trained model. Despite the wide use, it is not yet understood to which extent the obtained interpretation results can be trusted and how accurate they can reflect the model decisions. In order to fill this research gap, we conduct a study to evaluate different deep interpretation techniques quantitatively on EEG datasets. The results reveal the importance of selecting a proper interpretation technique as the initial step. In addition, we also find that the quality of the interpretation results is inconsistent for individual samples despite when a method with an overall good performance is used. Many factors, including model structure and dataset types, could potentially affect the quality of the interpretation results. Based on the observations, we propose a set of procedures that allow the interpretation results to be presented in an understandable and trusted way. We illustrate the usefulness of our method for EEG-based BCI with instances selected from different scenarios.**

## I. Introduction

A brain-computer interface (BCI) builds a direct communication pathway between the brain and external systems. Among the various neuroimaging techniques for BCIs, electroencephalography (EEG) is the most widely used method due to its noninvasiveness, affordability, and convenience. As one of the most powerful techniques to decode EEG signals, deep learning can automatically capture essential characteristics from a large volume of data by optimizing its parameters through back-propagation and stochastic gradient descent (SGD). It is reported that deep learning has achieved better performance than conventional methods in many BCI domains such as identifying attentive mental state [1], movement-related cortical potential recognition [2], detection of driver drowsiness [3], etc. Despite the success, deep learning has its major drawback of lacking transparency behind its behaviors, which could raise potential concerns of end users on adoption of BCIs.

In recent years, many efforts have been made to interpret the decisions of deep learning models with application to image and text classification tasks. This is commonly done by generating a heatmap indicating to which extent each pixel of the input contributes to the final classification for a trained model. For the context of EEG-based BCI, the technique can reveal how different components dwelling locally in EEG, e.g., signals generated from different cortical sources, sensor noise, electromyography (EMG), eye movements and eye blinks activities, will affect the classification [3-5]. It is thus possible to know whether the model has learned neurologically meaningful features, or the decisions are influenced largely by class-discriminative artifacts from the data, so that the process of improving the models towards better performance and reliability can be facilitated.

Deep learning interpretability has received wide attention in the field of EEG-based BCI [3-8]. Despite the wide use, it is neither well understood to which extent the obtained interpretation results can be trusted and how accurate they can reflect the model decisions nor clearly explained in existing literature why a specific interpretation technique is chosen over others. These observations raise concern on biased conclusions that are made based on misinterpretation of the model decisions. In order to fill this research gap, we conduct a study to evaluate different deep interpretation techniques for EEG-based BCI and explore the best practice of utilizing the techniques. To summarize, the paper makes contributions in the following aspects:

- By far as we know, this is the first comprehensive evaluation on deep learning interpretation techniques across different models and datasets for EEG-based BCI. It provides insights on how seven well-known interpretation techniques, including saliency map [9], deconvolution [10], guided backpropagation [11], gradient×input [12], integrated gradient [13], LRP [14] and DeepLIFT [12], behave under different conditions.
- Based on the evaluation results, we propose a set of procedures that allow sample-wise interpretation to be presented in an understandable and trusted way. We illustrate the usefulness of our method for EEG-based BCI

* This work was partially supported by the STI2030-Major Projects 2021ZD0200201, the National Natural Science Foundation of China (Grant No. 62201519), and Scientific Projects of Zhejiang Lab (No. 111012-AA2301, No. 2022KI0AC02, No. 2022ND0AN01).

*Corresponding author: Zhaoxiang Wang (email: wangzhaoxiang@zhejianglab.com); Ruilin Li: (RUILIN001@e.ntu.edu.sg)

[1] Research Center for Augmented Intelligence, Research Institute of Artificial Intelligence, Zhejiang Lab, Hangzhou, China

[2] Nanyang Technological University, Singapore

[3] Brainnetome Center, Institute of Automation, Chinese Academy of Sciences, Beijing 100190, China

with instances selected from different scenarios.

- We make the source codes that implement our method in this paper publicly available from [15]. This will allow other researchers from this field to conduct a quick test of their models or datasets in order to understand how the classification is influenced by different kinds of components in EEG signals.

In the following part of the paper, state-of-the-art interpretation techniques and their current applications to EEG signal classification are reviewed in Section II. Datasets and models are prepared in Section III. Deep learning interpretation techniques are selected and evaluated in Section IV. Sample-wise interpretation results are analyzed in Section V, which is followed by an extensive study on the applications in Section VI. The discussion is presented in Section VII and conclusions are made in Section VIII.

## II. Related work

### A. Deep Learning Interpretation techniques

In the field of deep learning interpretability, many techniques have been proposed to interpret deep learning models by generating a contribution map (or alternatively called "relevance" or "attribution" map [16]). Each value in the contribution map indicates the importance of the corresponding pixel (or sampling point) of the input sample to the final decision of the model. Existing interpretation techniques majorly fall into two categories – backpropagation-based methods and perturbation-based methods.

Backpropagation-based methods generate the contribution map through a single or several forward and backward passes through the network. The saliency map method [9] uses a direct way to estimate the contribution map by calculating the absolute values of gradients back-propagated from the target output. It reflects how much the target output will change when the input is perturbed locally. Zeiler and Fergus [10] proposed the deconvolution method, which modifies the back-propagation rule in the rectified linear units (ReLUs) layer – the backward gradients are zeroed out if their values are negative. By combining these two approaches, Springenberg et al. [11] proposed the guided backpropagation method which zeros out the gradients at the ReLU layer during back-propagation when either their values or values of inputs in the forward pass are negative. The gradient $\times$ input method [12] multiplies the (signed) partial derivatives with the input sample itself. Sundararajan et al. [13] proposed the integrated gradient method, where the average gradient is computed by varying the input along a linear path from a baseline. Bach et al. [14] proposed the layer-wise relevance propagation (LRP) method which redistributes the activation values at the target neuron to the neurons connected to it according to their contributions. The redistribution process continues layer by layer until the input layer is reached. Shrikumar et al. [12] proposed the DeepLIFT method, which requires running twice forward pass with the input sample and the baseline. Similar to the LRP method, each neuron is assigned with a contribution score in a top-down manner according to the difference of activations obtained from the twice forward passes.

The perturbation-based methods only focus on the change of output by perturbation of input, while treating the network as a black box. Specifically, such methods compute the difference of output when removing, masking or altering the input sample. Zeiler and Fergus [10] proposed the occlusion sensitivity method which sweeps a 'grey patch' to occlude different parts of an input image and observe how the prediction changes. Similar to the method, Petsiuk et al. [17] proposed to use binary random masks to perturb the image and distribute the contribution scores among the pixels. Zintgraf et al. [18] proposed the prediction difference analysis method. They calculated the difference of a prediction by marginalizing over each feature (or pixel). Fong and Vedaldi [19] proposed to use a soft mask with continuous values to preserve discriminative regions for classification. The soft mask is optimized with various regularizations to suppress artifacts. The method was further improved by Yuan et al. [20] by using a discrete mask optimized with the generative adversarial network.

These interpretation techniques have been previously evaluated on both real [21] and synthetic [22] image datasets, as well as synthetic time-series datasets [23]. However, they have not yet been systematically evaluated on EEG datasets. In this paper, we design metrics to evaluate how accurately these techniques can interpret the deep learning models designed for EEG-based BCI.

### B. Deep learning Interpretability for EEG-based BCI

For EEG-based BCI, deep learning interpretability can reveal how different factors contained in EEG influence the model decisions. For example, Bang et al. [6] compared sample-wise interpretation by the LRP method between two subjects and analyzed the potential reasons that lead to the worse performance on one of them. The LRP method was also used by Sturn et al. [7] to analyze the deep learning model designed for a motor imagery task. They attributed the factors leading to wrong classification to artifacts from visual activity and of eye movements, which dwell in EEG channels from occipital and frontal regions. Özdenizci et al. [24] proposed to use an adversarial inference approach to learn stable features from EEG across different subjects. By interpreting the results with the LRP method, they showed their proposed method allowed the model to focus on neurophysiological features in EEG while be less affected by artifacts from occipital electrodes. Cui et al. [3] used the Class Activation Map (CAM) method [8] to analyze individual classifications of single-channel EEG signals collected from a sustained driving task. They found the model had learned to identify neurophysiological features, such as Alpha spindles and Theta bursts, as well as features resulted from electromyography (EMG) activities, as evidence to distinguish between drowsy and alert EEG signals. In another work, Cui et al. [5] proposed a novel interpretation technique by taking advantage of hidden states output by the long short-term memory (LSTM) layer to interpret the CNN-LSTM model designed for driver drowsiness recognition from single-channel EEG. The same group of authors recently reported a novel interpretation technique [5] based on combination of the CAM method [8] and the CNN-Fixation methods [25] for multi-channel EEG signal classification and discovered stable features across different subjects for the task of driver drowsiness recognition. With the interpretation technique, they also analyzed reasons behind some wrongly classified samples.

Despite the progress, it is yet not understood to which extent the interpretation results can be trusted and how accurate they can reflect the model decisions. It is also not well explained in existing work why a specific interpretation technique is chosen over others. This research gap motivates us to conduct quantitative evaluations and comparisons of these interpretation techniques on deep learning models designed for mental state recognition from EEG signals.

## III. Preparation on datasets and models

### A. Dataset

We select three public datasets belonging to active, reactive and passive BCI domains, respectively, for this study. These datasets have been widely used for development of deep learning models.

*Dataset 1: Sensory motor rhythm (SMR).*

Motor imagery (MI) is an active BCI paradigm that decodes commends of users when they are imagining the movements of body parts [26]. It is reflected in EEG as desynchronization of sensorimotor rhythm (SMR) over the corresponding sensorimotor cortex areas. The EEG dataset comes from BCI Competition IV Dataset 2A [27]. It consists of EEG data collected from 9 subjects conducting four different motor imagery tasks, which are the imagination of moving left hand (class 1), right hand (class 2), both feet (class 3), and tongue (class 4). There are two sessions of the experiment conducted on different days for each subject. Each session consists of 6 runs separated by short breaks and each run consists of 48 trails (12 for each imaginary task). Therefore, there are in total 288 trails of a session for each subject.

The EEG data were collected from 22 channels with a sampling rate of 250 Hz. They were bandpass filtered to 0.5 Hz-100 Hz, and further processed by a 50 Hz notch filter to suppress line noise. We followed the practice in [2] to down-sample the signals to 128 Hz and extracted the EEG samples for each trail from 0.5 to 2.5 seconds after the cue appeared. The dimension of each sample is therefore 22 (channel) × 254 (sample points).

*Dataset 2: Feedback error-related negativity (ERN)*

Feedback error-related negativity (ERN) refers to the amplitude change of EEG, featured by as a negative error component and a positive component, after a subject receives an erroneous visual feedback [28]. In the experiment, ERN was induced by a P300 speller task, which is a passive BCI paradigm for selecting items displayed on the screen by detecting P300 response from EEG signals. The experiment consists of five sessions. Each of the first four sessions contains 12 tasks of 5-letter word spelling, while the last session contains 20 tasks. For each subject there are 4 (sessions) × 12(tasks) × 5(letters) + 1(session) × 20 (tasks) × 5(letters) = 340 trails. 26 subjects (13 males) aged between 20 and 37 participated in the experiment. Their EEG data were recorded at 600 Hz by with 56 passive Ag/AgCl EEG sensors (VSM-CTF compatible system) placed according to a 10-20 system. The authors down-sampled the EEG signals to 200 Hz and divided them into training data (from 16 subjects) and testing data (from 10 subjects). The dataset has been made public from 'BCI Challenge' hosted by Kaggle [29].

We selected 32 channels out of 56 channels in our study by following the practice described in [28]. Next, we processed the data according to steps used in [2] by band-pass filtering the signals to 1–40 Hz and extracting a sample from each trail from 0-1.25 seconds after the feedback was displayed. In this way, each sample has dimensions of 32(channels) × 200(sampling points).

*Dataset 3: Driver drowsiness recognition*

This dataset was built with EEG data collected from a sustained-driving experiment [30]. The subjects were required to drive a car in a virtual experiment and respond quickly to randomly introduced lane-departure events that drifted the car away from the center of the road. Their reaction time was recorded to reflect the level of drowsiness. 27 subjects (aged from 22 to 28) participated in the experiment. The EEG signals was recorded at 600 Hz with 30 electrodes, and band-pass filtered to 1-50 Hz followed by artifact rejection.

We use pre-processed version of the dataset described in [3] for this study. Specifically, the EEG data were down-sampled to 128 Hz. A 3-second length sample prior to each car deviation events was extracted. The dimension of each sample is 30 (channel) × 384 (sample points). The samples were labeled with 'alert' and 'drowsy' according to their corresponding global and local reaction time, which were defined in [31]. The samples were further balanced for each subject and class. The final dataset contains 2022 samples in total from 11 different subjects.

### B. Models and implementation

In this study, we select two benchmark deep learning models for the test. The first model is a shallow CNN named "EEGNet", which was proposed by Lawhern et al. [2]. The model was tested on different active and reactive BCI paradigms [2]. The second model named "InterpretableCNN" was proposed by Cui et al. [5] in their recent work for driver drowsiness recognition. The model has a compact structure with only seven layers. Selection of the model for this study is motivated by its superior performance over both conventional methods and other state-of-the-art deep learning model including EEGNet on the passive BCI domain [5].

The cross-subject paradigm was carried out on the three paradigms, in order to encourage the models to derive stable EEG features across different subject. For Dataset 1, we followed the procedures described in [27] by splitting the data collected from the first and second sessions into training and testing sets. For each time, the data from one subject collected from the second session are used as the test set, while the data of all the other subjects collected in the first session are used as the training set. The process was iterated until every subject served once as the test subject. For Dataset 2, we followed the original division of the dataset [29] by using the data from 16 subjects as the training set and the data from the other 10 subjects as the testing set. For Dataset 3, we followed the practice in [5] by conducting leave-one-subject-out cross-subject evaluation on the models.

We set batch size as 50 and used default parameters of the Adam method [32] ($\eta$ = 0.001, $\beta_1$ = 0.9, $\beta_2$ = 0.999) for optimization. Considering Dataset 2 contains imbalanced samples, we applied the weights of 1 and 0.41 (which is inverse proportion of the training data) to the "error" and "correct" classes, respectively, in the loss function. Considering the neural networks are stochastic, we repeated evaluation of each model on each test subject for 10 times. In each evaluation, we randomized the network parameters and trained the models from 1 to 50 epochs and selected the best epoch for the three datasets, respectively.

## IV. Evaluation on deep learning interpretability

### A. Interpretation techniques

#### 1) Statement of the problem

Formally, suppose an EEG sample $X \in R^{N \times T}$ ($N$ is the number of channels and $T$ is the sample length) is predicted with label $c$. The task is to generate a contribution map $S_c \in R^{N \times T}$, which assigns a score $S_c(i,j)$ $(1 \le i \le N, 1 \le j \le T)$ to each sampling point $X(i,j)$ indicating its contribution to the classification.

By averaging $S_c$ over the temporal dimension, we can obtain a mean contribution map $\overline{S_c} \in R^N$, reflecting the average contribution of each EEG channel to the final classification. By interpolating $\overline{S_c}$ over the whole scalp area, we can obtain a topographic map that reveals the source of signals that contain important features. $\overline{S_c}$ has been widely used in existing work [5, 6, 24] to interpret the classification results. In this paper, we name $\overline{S_c}$ as "channel contribution map". $S_c$ is called "contribution map" or alternatively "sample contribution map", referring to the map generated for the whole sample.

#### 2) Interpretation techniques and implementation

We select seven widely used interpretation techniques for the test and they are the saliency map [9], deconvolution [10], guided backpropagation [11], gradient×input [12], integrated gradient [13], LRP [14] and DeepLIFT [12]. In comparison to the other methods, the selected ones have the following advantages to be implemented for EEG-based BCI:

- The methods can be applied to deep learning models with different structures.
- The selected interpretation techniques are free of adjustable parameters to be fine-tuned.
- The selected methods are computationally efficient.
- The methods are free of parameters to be randomized for initialization, so that the results are reproducible.

The selected interpretation methods were implemented for EEGNet and InterpretableCNN with the Pytorch library. We used the input sample with zero entries as baselines [16] for integrated gradient and DeepLIFT. We computed the average of gradients generated from the path between the baseline and the input with 100 steps for the integrated gradient method. We implemented the LRP with $\epsilon$-rule and DeepLIFT with rescale-rule by modifying the gradient flow in the nonlinear activation layers (the ReLU activation layer for InterpretableCNN and three ELU activation layers for EEGNet) with the method proposed by Ancona et al. [16]. LRP is equivalent to gradient × input for InterpretableCNN as the model only has ReLU activation for the nonlinear layer [16]. In order to remove the influence of other samples from the same batch on the interpretation results, we made the batch normalization layers behave linearly by fixing the parameters of batch mean and standard deviation during backpropagation. The parameters were obtained from an additional forward pass of the tested batch data.

### B. Evaluation metrics

#### 1) Sensitivity test

Our first test was inspired by the sensitivity-*n* test proposed by Ancona et al. [16]. In the method, they randomly perturbed *n* pixels (by setting their values to zeros) from the input sample and observed change of output. Ideally, the sum contribution of the *n* points is proportional to the change of model output score of the predicted class. They varied *n* from 1 pixel to about 80% pixels and calculated Pearson correlation coefficient (PCC) *r* for each *n* as the quality metric of the contribution map.

Our test is different in the aspect that we perturbed the input sample locally by small patches with fixed length *n* instead of *n* random points from different parts of the sample. In this way, we can evaluate local accuracy of the contribution map while introducing less high-frequency noise to EEG signals. We limit the patch size *n* to 0.1-0.5 of the sample length, which correspond to 0.45%-2.27%, 0.31%-1.56%, and 0.33%-1.67% of the sample size for Dataset 1, 2 and 3, respectively. We assume the perturbation will not significantly drift the sample away from its original distribution. For each *n*, we randomly perturbed the sample 100 times and calculate the correlation coefficient as a quality metric of the contribution map.

The channel contribution map was evaluated with a single correlation coefficient, which is obtained by perturbing each channel once.

#### 2) Deletion test

The deletion test proposed in [17] is used in this study. In this test, we ranked the sampling points of the input sample in a descending order according to their scores in the contribution map. By varying *n* from 1%, 2%, …, 100% of the sample size, we calculated the probabilities of the predicted class when the first *n* points were removed from the sample by setting their values to zeros. A sharp drop of the probability on the predicted class, or alternatively a small area under the probability curve (as a function of *n*) is indicative of a high-quality contribution map.

We performed the deletion test for the channel contribution map in a similar way – each time we removed a channel by setting them to zeros and calculated the probability of the predicted class.

### C. Test settings

Dataset 1 contains EEG data collected from 9 subjects. Each subject has 288 training samples and 288 testing samples with balanced 4 classes. We randomly selected 25 samples for each class from the test samples of each subject. In this way, we have in total 9 (subjects) × 4 (classes) ×25 (samples) = 900 samples for evaluating the interpretation results.

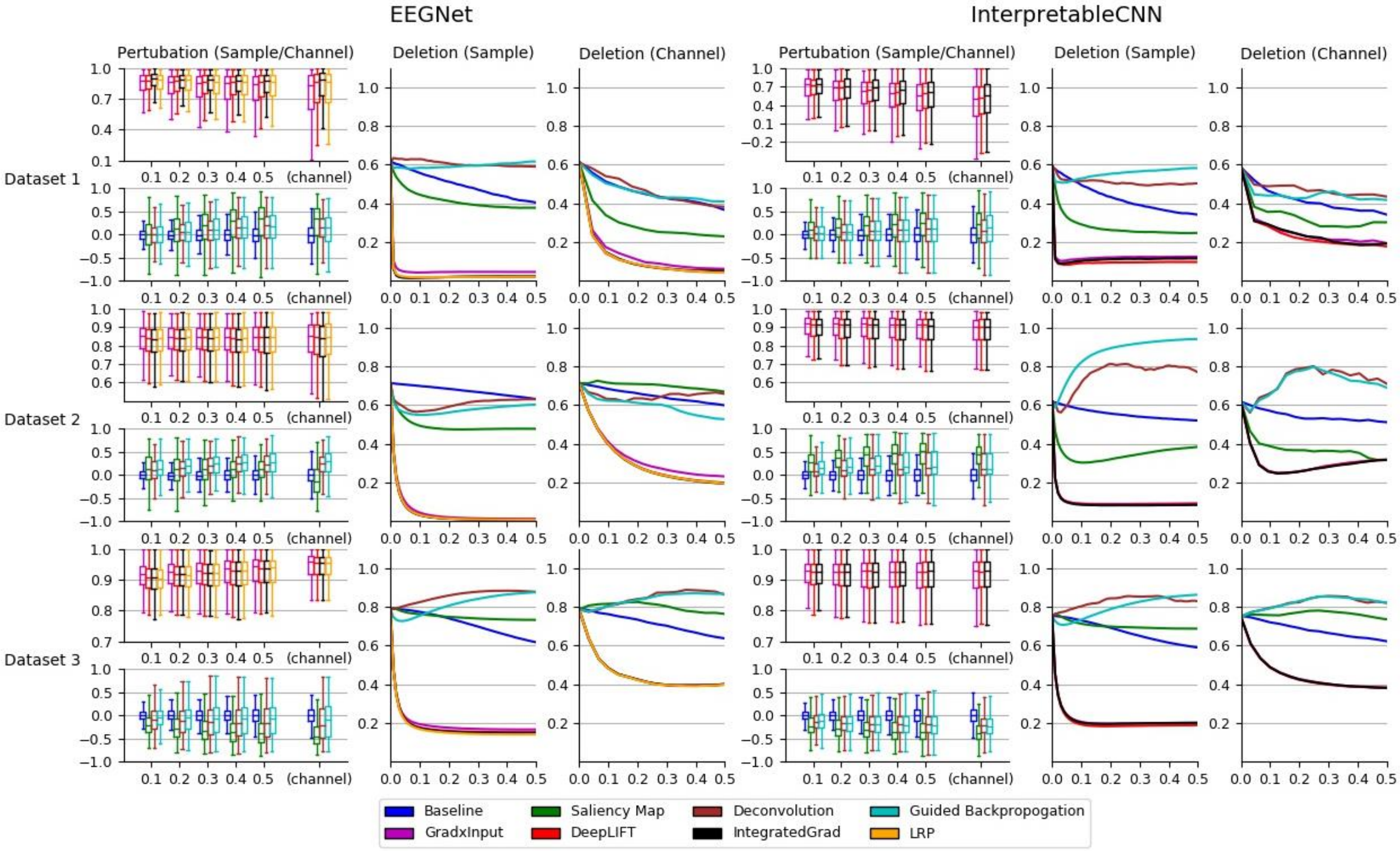

Fig. 1. Evaluation results of the interpretation techniques for InterpretableCNN and EEGNet on the three datasets. The results for EEGNet and InterpretableCNN are displayed in columns 1-3 and columns 4-6, respectively. The results for the three datasets are displayed in the three rows, respectively. The sensitivity test results are shown in column 1 and 4. We display the boxplot of the correlation coefficients *r*, which is in the range of -1 to 1 (1 represents a perfect correlation). The sensitivity test results for both the sample and channel contribution maps are displayed in the same sub-figure while different interpretation techniques are grouped and separately displayed in two sub-figures. For InterpretableCNN, results for the LRP method are not shown since they are identical to that obtained with the gradient×input method. The deletion test results for the sample contribution map are shown in columns 2 and 5, while the results for the channel contribution map are shown in column 3 and 6. For the deletion test, we show the average probabilities of the predicted class against the fraction of 0-0.5 of the sample size.

Dataset 2 contains EEG data collected from 16 training subjects and 10 testing subjects. Each subject has samples with unbalanced labels. We selected 100 samples from each test subject and thus have in total 10 (subjects) ×100 (samples) = 1000 samples for evaluation. For the test subjects 1-6, 9 and 10, we randomly selected 50 samples of each class from each subject. For subjects 7 and 8 with less of 50 samples for the class of error feedback, we used all the samples from this class and randomly selected the rest samples from the class of correct feedback.

Dataset 3 contains EEG data collected from 11 subjects. We randomly selected 50 samples of each class from each subject, and thus have in total 11 (subjects) × 2 (classes) × 50 (samples) = 1100 samples for evaluation.

For each sample, we generated a contribution map with random values as baseline. The corresponding channel contribution map is obtained by averaging the sample contribution map over the temporal dimension.

### *D. Results*

#### *1) Performance of different interpretation techniques*

As it can be seen from the results displayed in Fig. 1, the interpretation techniques fall into two groups by their performance in the tests. The first group of methods, consisting of gradient×input, DeepLIFT, integrated gradient, and LRP, have similar and better performance than the baseline method, which uses randomly generated contribution maps, while the second group of methods consisting of saliency map, deconvolution and guided backpropagation fail to outperform the baseline method in most conditions. Specifically, in the sensitivity tests the medians of correlation coefficients of the first group of methods range from 0.4 to 1 under different the conditions, while the medians of the second group of methods are mostly around 0 under different the conditions, which fail to outperform the baseline method. In the deletion tests, the first group of methods reach a low probability on the predicted class (below 0.2 under all the conditions) when less than 0.1 portion of the total data are removed under different conditions, indicating a small portion of features that contribute most to the classification have been successfully localized. However, the second group of methods mostly have large AUC (Area under the Curve), indicating the most important features learned by the deep learning models for classification are not accurately localized.

The study justifies the usefulness of the gradient×input, DeepLIFT, integrated gradient, and LRP methods for interpreting deep learning models designed for classifying EEG signals. It can be observed that all these four methods have similar performance in both tests under different conditions. The reason could be attributed to the limited number of non-linear layers, where lie the major differences of how these methods deal with backpropagation, of both the deep learning models selected for the study. The results also reveal that the methods of saliency map, deconvolution and guided backpropagation may not be suitable for interpreting deep learning models designed for EEG signal classification. The observation raises concern on potential misinterpretation of the model decisions in existing literature work, e.g., [33], where any of these methods are used.

In the following part of this paper, we focus on methods of gradient×input, DeepLIFT, integrated gradient, and LRP. We use them alternatively for interpreting sample-wise classification results.

*2) Quality of individual interpretation*

Despite when we focus only on interpretation results produced with best available methods of gradient×input, DeepLIFT, integrated gradient, and LRP, there still exists a large variation on the quality of individual samples under certain circumstances. As it can be seen in Fig. 1, the correlation coefficients have a wide range for Dataset 1 and it increase along with the size of the perturbation patch. When size reaches 0.5 of the sample length, the correlation coefficients range from 0.4 to 1 for EEGNet and -0.2 to 1 for InterpretableCNN, while in comparison the range falls stably within 0.75 to 1 for Dataset 3 under different conditions. The results reveal that the interpretation results generated for some samples are not meaningfully correlated locally with the model outputs. In addition, the model structure also has an impact on the quality of interpretation results. Specifically, for Dataset 2 the correlation coefficients range from 0.6 to 1 for EEGNet, while they range from around 0.7 to 1 for InterpretableCNN. The apparent difference on performance for different models can also be observed from the tests on Dataset 1.

To summarize, both the dataset type and model structure have impact on the quality of interpretation results, while this problem may not be perfectly solved by changing the interpretation technique. The individual interpretation results should therefore be cautiously treated, since they could be uninformative or even misleading for many samples. We further discuss how the individual interpretation results can be presented in an understandable and trusted way in Section V.

## V. Proposed Method for Sample-Wise Interpretation

*1) Enhancement of visualization*

The contribution maps are commonly visualized as heatmaps [8, 34] or colormaps [3-5] after normalization. However, the colormaps produced in a such way tend to be too noisy to be interpretable for the backpropagation-based methods investigated in this paper. We show a concrete sample in Fig. 2(a), in order to illustrate the problem we meet. The sample is obtained from Dataset 3. It is predicted correctly with the label of “drowsy” by InterpretableCNN with probability of 1, indicating that features strongly correlated with the drowsiness have been identified by the model. The sample contains apparent alpha spindles in its first half part (around 0-1.5 seconds), which were found to be a strong signal of drowsiness [3]. The contribution map is obtained with the grad × input method, and it is visualized in Fig. 2(a) as a colormap directly after normalization. However, it is difficult to observe any meaningful pattern from the contribution map, as it is corrupted by the heavy high-frequency noise. Actually, we would expect to observe distinguishable features from FCZ, CZ, CPZ and FT7 channels, as which are highlighted in the topographic map.

In order to enhance the visualization, we propose to conduct two additional steps consisting of thresholding and smoothing after normalization. Specifically, we manually set a threshold and remove the unnecessary information from the normalized contribution map below this threshold, in order to reduce the abundant information contained in the sample. After that, we conduct smoothing by moving an average window in order to make the features distinguishable. Coming back to the sample shown in Fig. 2(a), we preformed the proposed processing steps with the thresholds of 2 and 1 for the sample and channel contribution maps, respectively, and the smoothing window with size of 5. As it can be seen in Fig. 2(b), the visualization is apparently improved – the alpha spindle features are clearly visible from the channels of FCZ, CZ, CPZ and FT7, which is consistent with the information revealed from the channel contribution map.

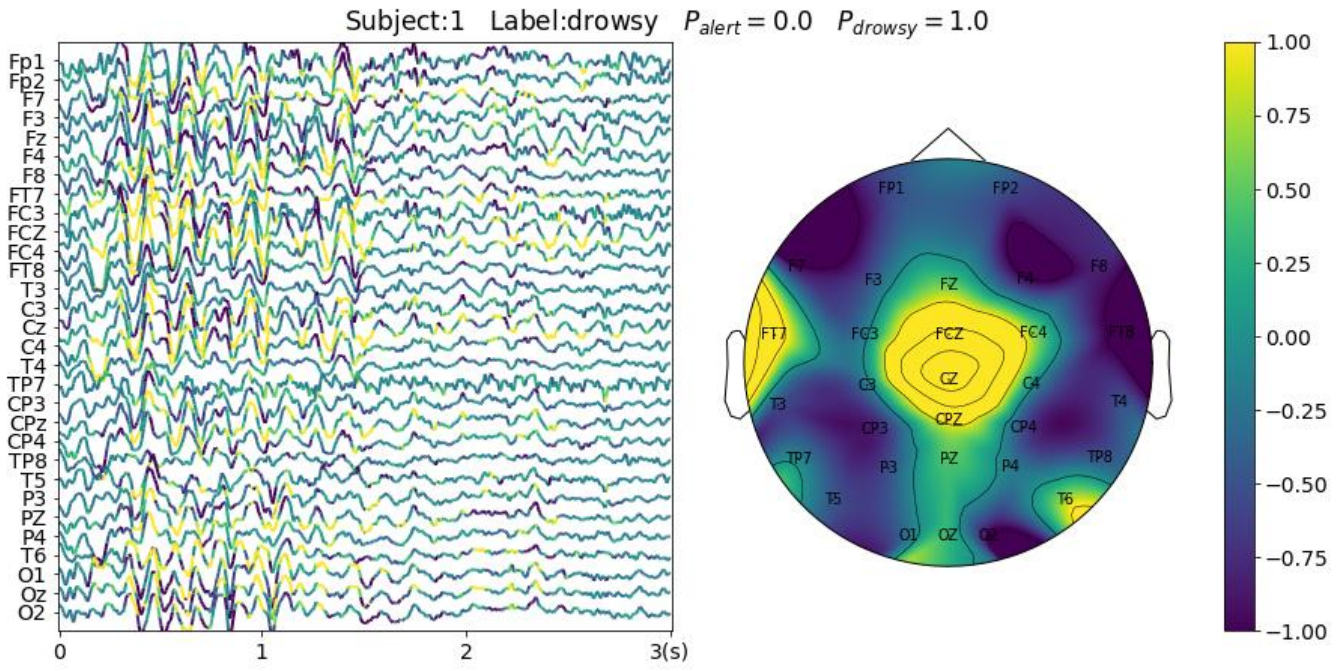

(a)

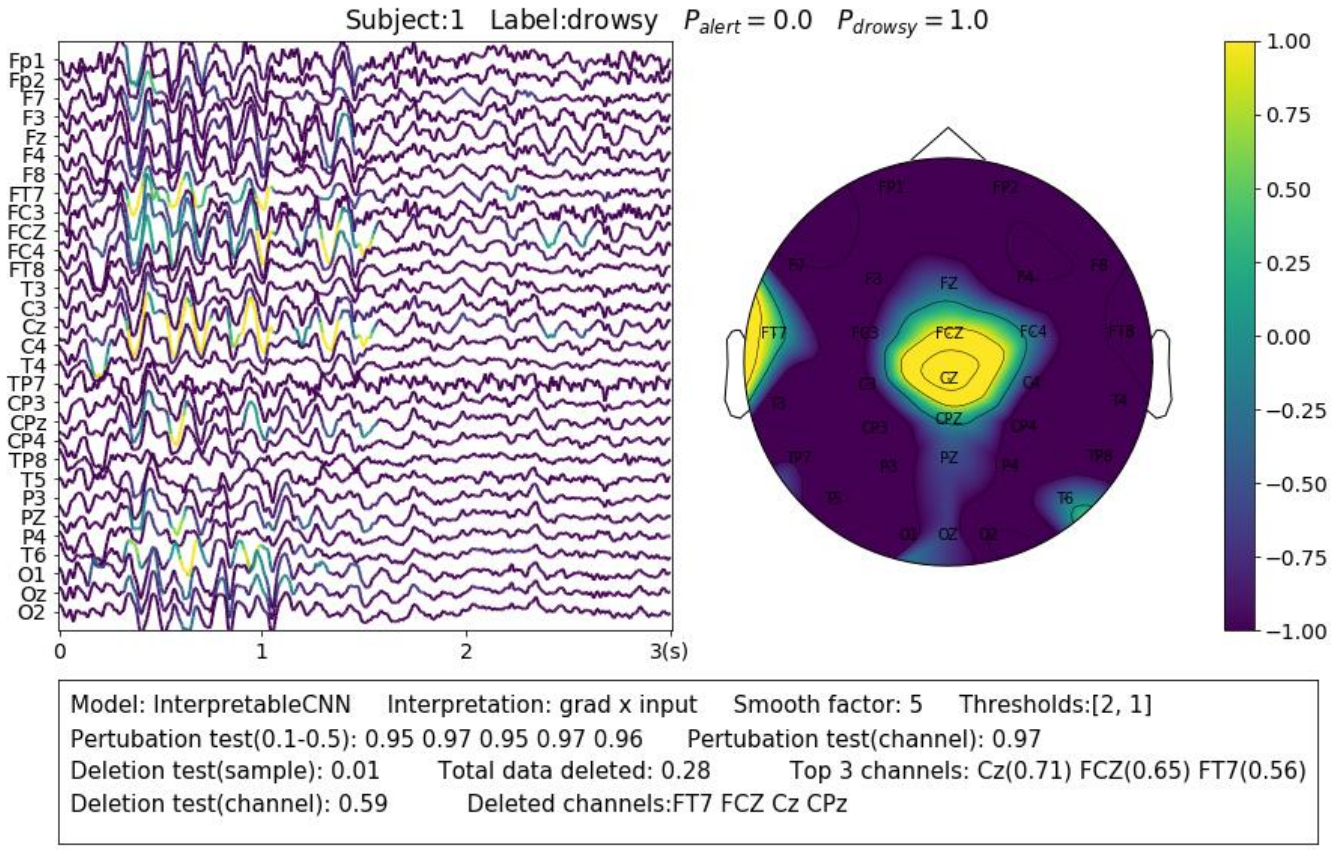

(b)

Fig. 2. Comparison of the contribution maps visualized (a) simply after normalization and (b) after processing with the proposed pipeline for a sample selected from Dataset 3. The subject ID, ground truth label, probabilities of both classes are shown on the top of each sub-figures. The contribution scores are converted to colors overlaid on the signals. An evaluation report is generated and attached below the instance in (b). In the report, the deep learning model, the interpretation technique, the smoothing window size and the thresholds (for the sample and channel contribution maps) are displayed in the first line. The perturbation testing results for the sample and channel contribution maps are displayed in the second line. The deletion test results for the sample and channel contribution maps are displayed in the third and fourth lines, respectively. The first item in the third line shows the new probability after deleting the highlighted parts of the signal. The second item shows the total amount (portion) of data deleted. The third item shows the top 3 channels that contain the most of the deleted data and the amount (portion) of deleted data from them. In the fourth line, the first item shows new probability after deletion of the channels listed in the second item.

*2) Generation of sample-wise evaluation*

As it has been discussed in the Section III D. 2), there exists a large variation on the quality of interpretation results for individual samples even when the best available interpretation technique is used. It is therefore important to conduct sample-wise evaluation and present the results along with the interpretation, since the contribution maps themselves cannot reflect how accurate the model decisions are interpreted. In this way, the interpretation results of low accuracies can be excluded so that misinterpretation of the model decisions can be to a large extent avoided.

To keep the consistency of the paper, we use the sensitivity and deletion tests as described in Section IV. B for the evaluation. The sensitivity test is conducted on the original contribution maps to reflect the best correlation achieved between the perturbed batches and the model output. The deletion test is conducted on the processed contribution map – we remove the highlighted areas (for the sample contribution map) or the channels (for the channel contribution map) and the report the probability output by the model on the predicted class. In this way, the deletion test results can be directly related to what is observed from the displayed colormap. We generate the evaluation report in the text box under each figure. The sample for illustration is shown in Fig. 2 (b).

We display another two samples in Fig. 3 for the purpose of illustrating the importance of sample-wise evaluation. By observation merely on the interpretation results of the sample shown in Fig. 3(b), we may draw the conclusion that the features recorded at around 0.75-1 second from CP3 channel contribute most to the wrong classification results. However, the evaluation results show that the local regions of the sample contribution map do not correlate well with the model output and removal of channel CP3 will not actually cause the prediction probability to drop. Without the evaluation, it will be easy to draw biased conclusions from the misleading interpretation results. Despite some factors that could potentially influence the quality of interpretation can be observed from the obtained results, e.g., the deep learning model structures and the different types of EEG features, it is yet not fully understood what actually lead to failure of interpretation for some samples, e.g., the one in Fig. 3(b). We leave further investigation on this topic to future work.

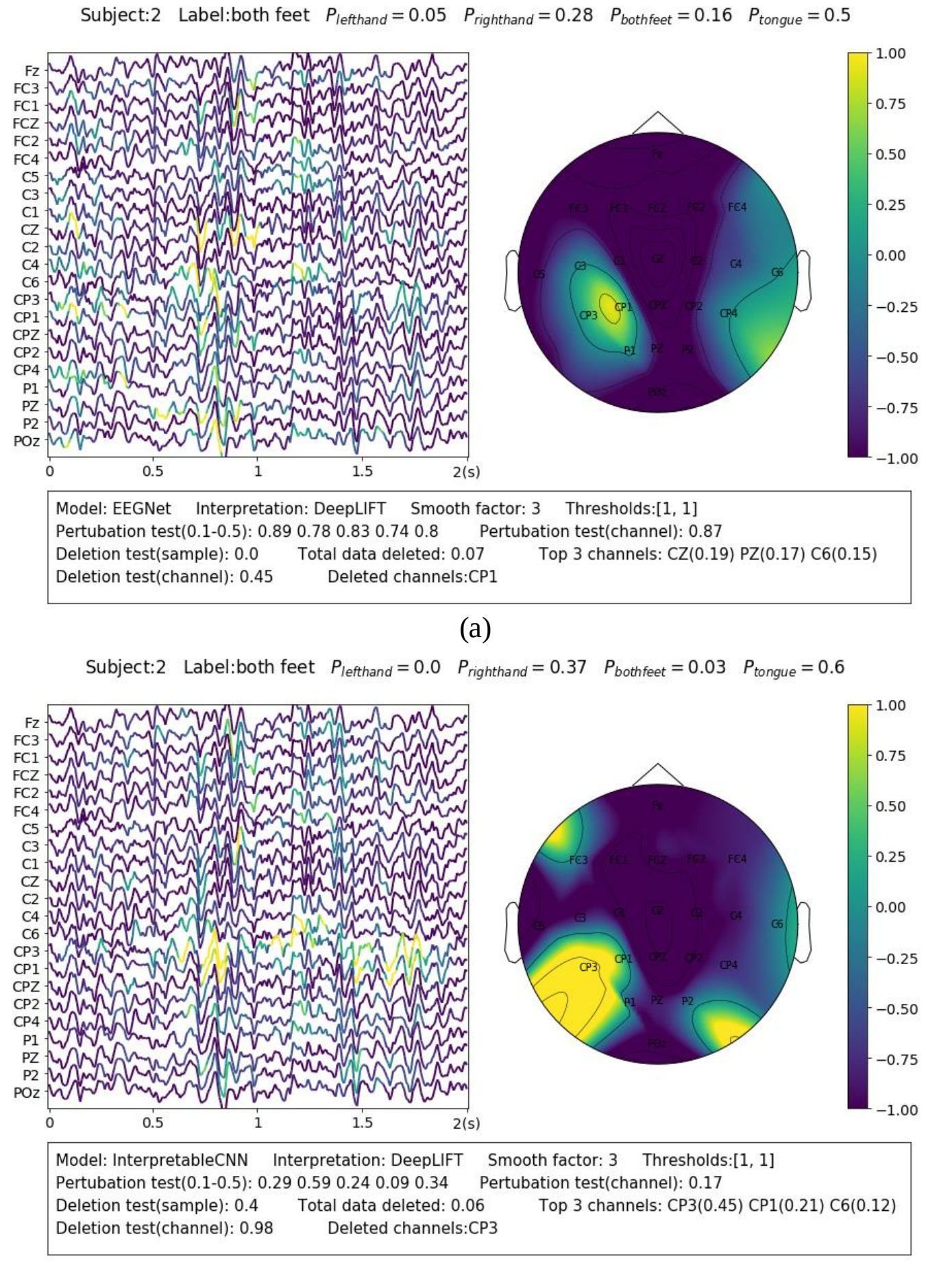

(b)

Fig. 3. Comparison of the interpretation results on a sample classified by (a) EEGNet and (b) InterpretableCNN. The contribution maps generated for the two samples display similar patterns, while the qualities of the interpretation results vary greatly as revealed by the evaluation report. For the sample shown in (a), the sample contribution map is well correlated with the model output with the correlation coefficients in the range of 0.74-0.89. Removal of the highlighted regions (taking up 0.07 of total data) of the first sample will cause the probability drop from 0.5 to 0.0, while removal of the channel (CP1) with the highest contribution will cause the probability drop slightly from 0.5 to 0.45. However, for second sample shown in (b), the local regions of the sample contribution map do not correlate well with the model output as the correlation coefficients fall in the range of 0.09-0.59. Removal of the most important channel (CP3) for the second sample will cause the probability, on the contrary, increase from 0.6 to 0.98. Therefore, the evaluation results allow us to confidently reject the interpretation shown in (b).

## VI. Application Scenarios

In this section, we extensively explore how we can benefit from interpreting deep learning models with the method proposed in this paper for different EEG-based BCIs. The applications are explored in two scenarios. In the first scenario, we visualize the neurophysiological features learned by the models from EEG for different datasets, which is an important step of model validation. In the second scenario, we show the advantages of using deep learning interpretability to discover different types of noise and artifacts in the datasets and discuss how the classification accuracy can be potentially improved based on the findings.

*1) Visualization of neurophysiological features*

Deriving insights into what the model has learned from the data is an initial step of model validation. The interpretation results allow us to know whether neurophysiological features have been learned from the data to distinguish different mental states. We have selected three representative samples from the three datasets, respectively, for the purpose of illustration.

For the SMR dataset (Dataset 1), the samples were collected while the subjects were imaging movements of different body parts. This reflects in EEG as event-related desynchronization (ERD) during imagination and event-related synchronization (ERS) after imagination of sensorimotor rhythm (SMR) or Mu rhythm (8-13Hz) over the corresponding sensorimotor cortex areas [35]. A representative sample is shown in Fig. 4(a). The sample is correctly predicted with label of "tongue movement" by EEGNet with probability of 0.95. From the channel contribution map, we can observe that the model has found important features from P1 and POZ channels, which are closest to the sensorimotor area of the tongue [36]. From the sample contribution map, it can be observed that the model has recognized the decreased amplitude sensorimotor rhythm (or ERD) at around 0-0.4 seconds from POz channel, as well as an instant burst of Mu spindles at around 0.4-0.6s as evidence for the prediction. The amplitude change of SMR reflects neuron activities resulted from the tongue imagination task in the corresponding sensorimotor area.

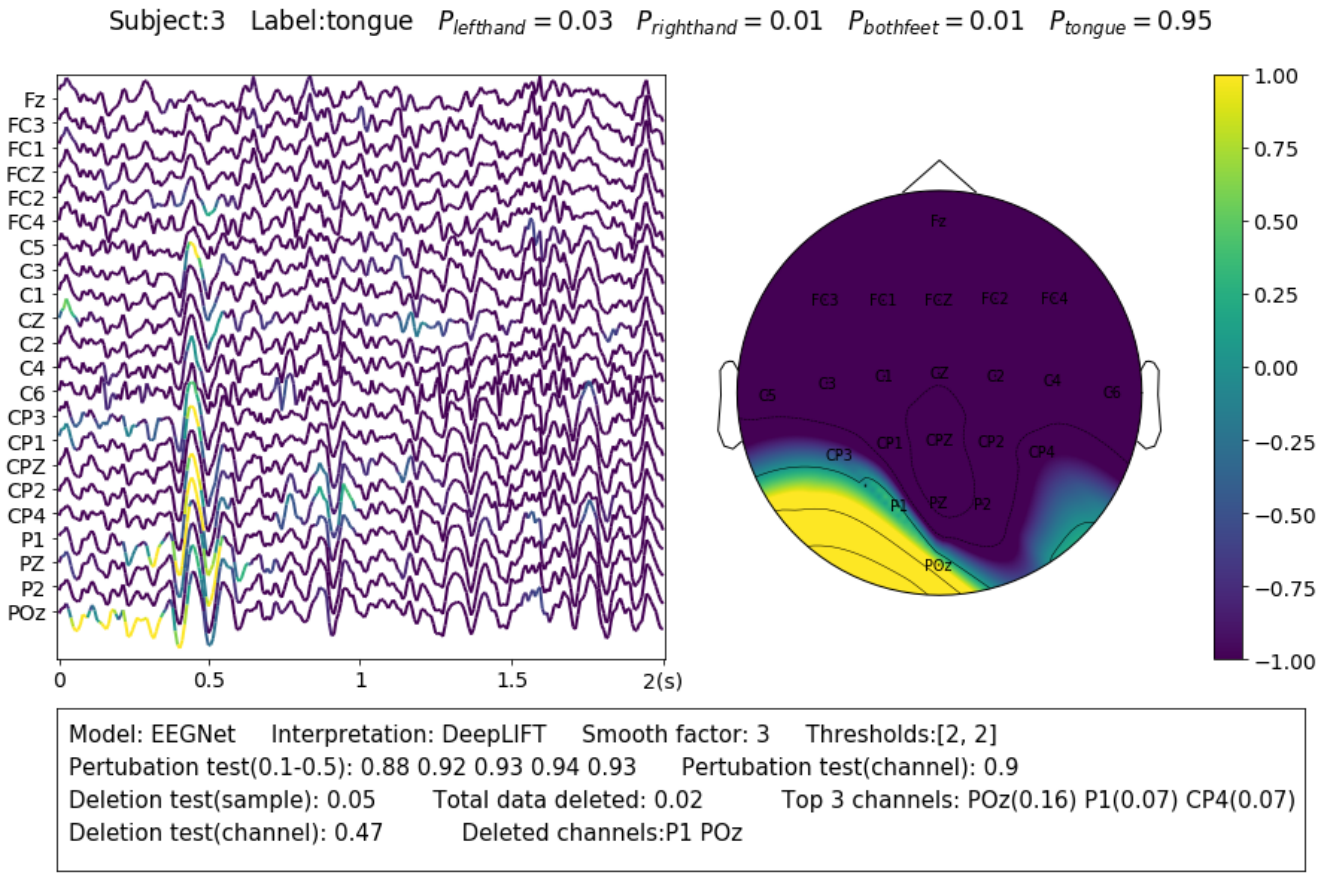

(a)

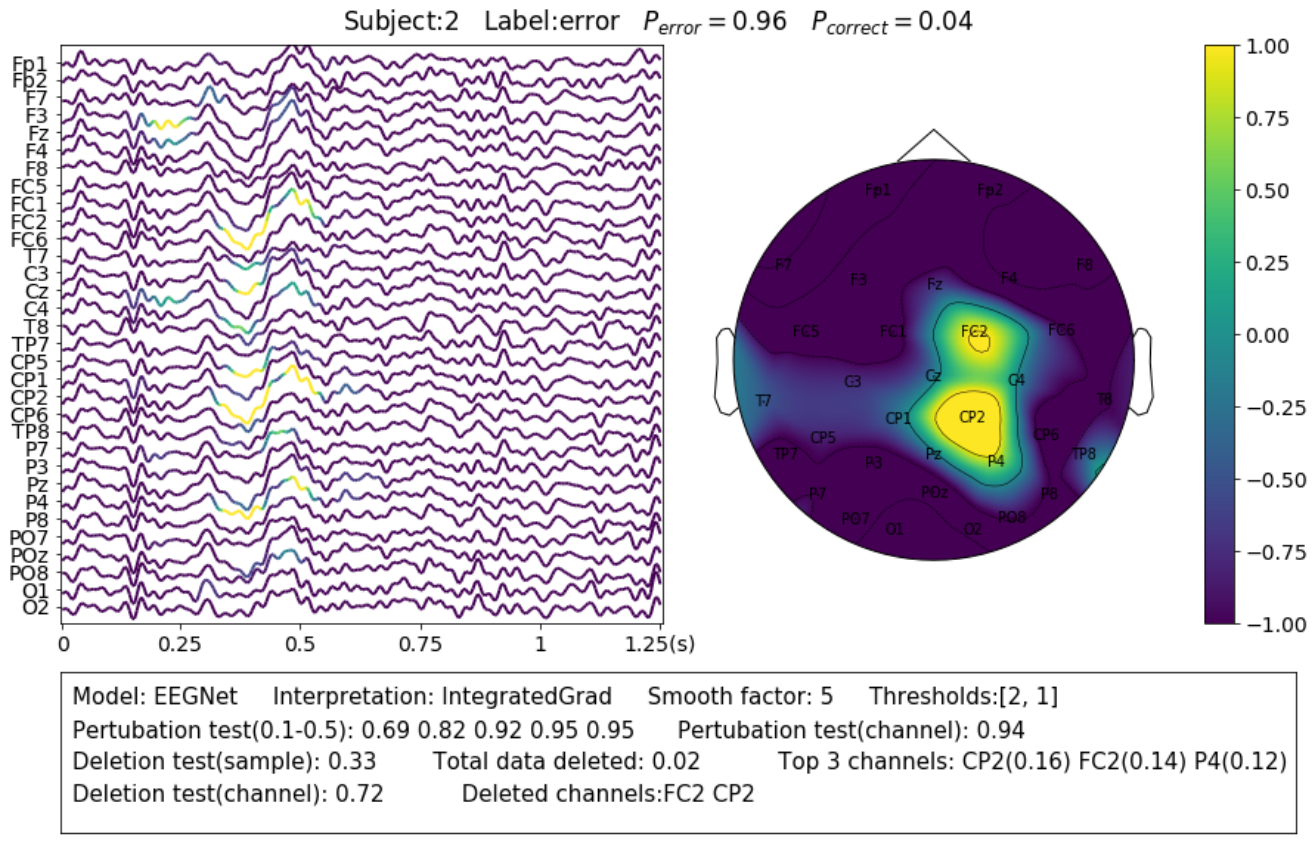

(b)

Fig. 4. Visualization of the learned neurophysiological features from two samples. The first sample (a) is from Dataset 3 and the second sample (b) is from Dataset 2.

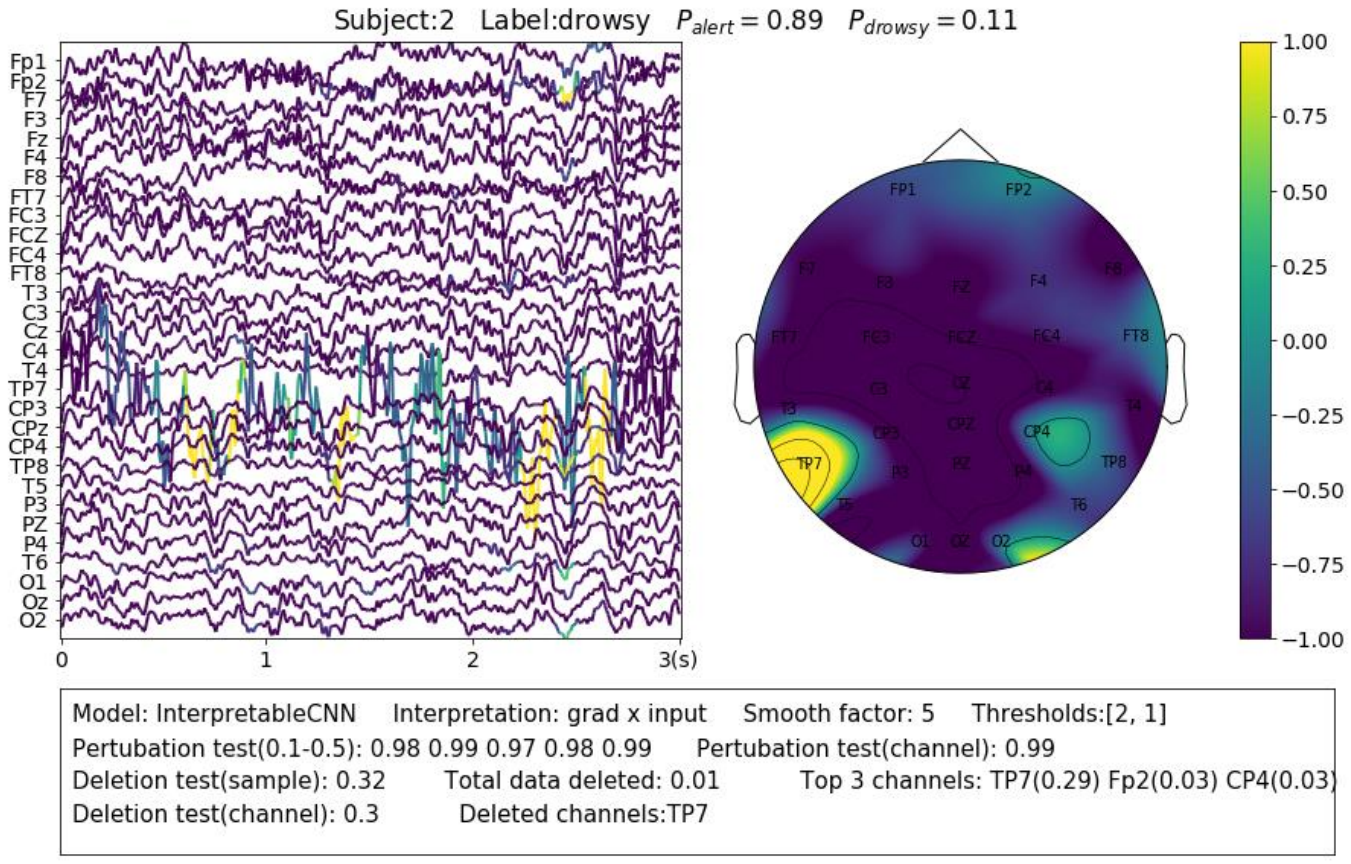

(a)

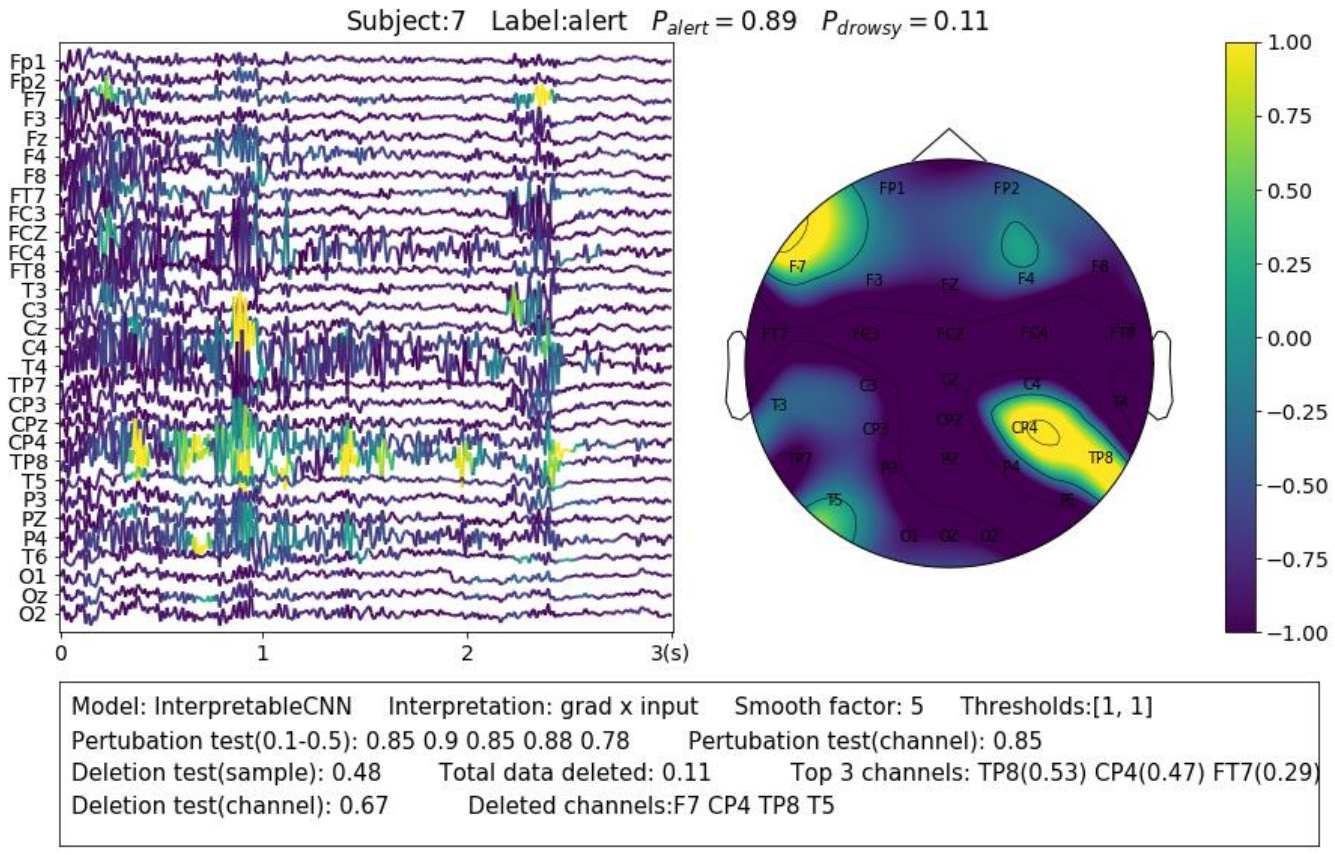

(b)

Fig. 5. Visualization of the interpretation results for selected samples containing different types of noise.

Feedback related negativity (FRN, or feedback ERN) refers to a specific kind of evoked responses produced by the brain when negative feedbacks are received from external stimuli. It is featured by a negativity peaking of EEG signals around 250 ms after feedback is presented [37]. For Dataset 2, FRN occurs when the subject receives error prediction of the letter displayed on the screen. As it can be seen in the example shown in Fig. 4(b), the model has identified the typical FRN feature, which has a negativity peaking followed by a positive peaking (see Figure 7 in [28]), at 0.25-0.5s after the feedback is presented. The sample is predicted with the "error" class with probability of 0.96. Removal of the highlighted areas (taking up 0.02 of total data) will cause the probability to drop from 0.96 to 0.33.

For Dataset 3, Alpha spindles, which are characterized with an arrow frequency peak within the alpha band and a low-frequency modulation envelope resulting in the typical 'waxing and waning' of the alpha rhythm [38], are the most notable features in EEG associated with drowsiness [5]. A typical sample is shown in Fig. 2(b). The sample is predicted correctly by InterpretableCNN with probability of 1.0. The Alpha spindle features have been identified by from several episodes of the signal majorly in the central cortical areas. Removal of the features (taking up 0.28 of total data) will cause the probability to drop from 1.0 to 0.28.

### 2) *Discriminating different noises and artifacts*

EEG recording is highly susceptible to various forms and sources of noise and artifacts. The sensor noise contained in EEG is one of the major reasons that affect the model decisions. The interpretation results allow us to understand how different kinds of sensor noise impact the model decisions. We selected two samples from Dataset 3 for illustration. The first sample shown in Fig. 5(a) contains apparent sensor noise in the TP7 channel. The model falsely identified several local regions of the sensor noise as evidence for decision. The significant amplitude changes of the signal in TP7 channel could be caused by loose conduct between the sensor and skin. Such kind of sensor noise that seriously affects the model decision should be cleaned from the data in the pre-processing phase. The second sample shown in Fig. 5(b) contains heavy high-frequency noise in several EEG channels, e.g., CP4. The model identifies several regions from the noise areas as evidence for classification. Actually, we have observed many samples containing similar noise in the dataset that are correctly classified with alert labels, which indicates that this kind of noise could have a strong relationship with the alert state. The noise is actually caused by electromyography (EMG) activities resulted from tension of scalp muscles. They usually dominate the wakeful EEG signals [39] and become the most apparent feature of alertness [3-5], while the cortical source Beta activities with very low amplitude [40] in wakeful EEG are not as easy to be distinguished.

Eye blinks and movements are another common source of artifacts in EEG signals. The interpretation results allow us to identify samples that contain such kind of artifacts and understand how they affect the model decision. For Dataset 2, we find there are many samples, similar to the case shown in Fig. 6(a), containing the eye blink artifacts while they are identified by the model as evidence of the "correct" class. The eye blinks could be a sign of relief after the subjects stare at the screen with high concentration in the P300 task. Such class-discriminative artifacts should be removed from the dataset and the model is expected to learn EEG features (e.g., ERP) generated from cortical sources instead. However, the case is on the contrary for Dataset 3, where the eye blink and movement features are overly cleaned. Actually the rapid eye blinks, as reflected in EEG a short-term pulse in the frontal channels (Fig. 6(b)), are indicators of the alert and wakeful state. They have been found by deep learning models as important features for classification [5]. However, the eye blink and movement features are overly cleaned in the pre-processing phase for Dataset 3, which causes the prediction accuracy to drop around 0.3 for Subject 2 in our test. We show in Fig. 6(b) an uncleaned sample from Subject 2 and it can be seen that the model has made the right prediction based on such kind of features. Removal of the features will cause the model to make the wrong prediction.

The observations above lead to the conclusion that different type of noise should be treated differently rather than indiscriminately removed from EEG signals. The noise and artifacts defined in one scenario could become important features in another scenario. Deep learning interpretability provides us with the advantage of understanding how they impact the model decisions so that we can take proper strategies accordingly to deal with them.

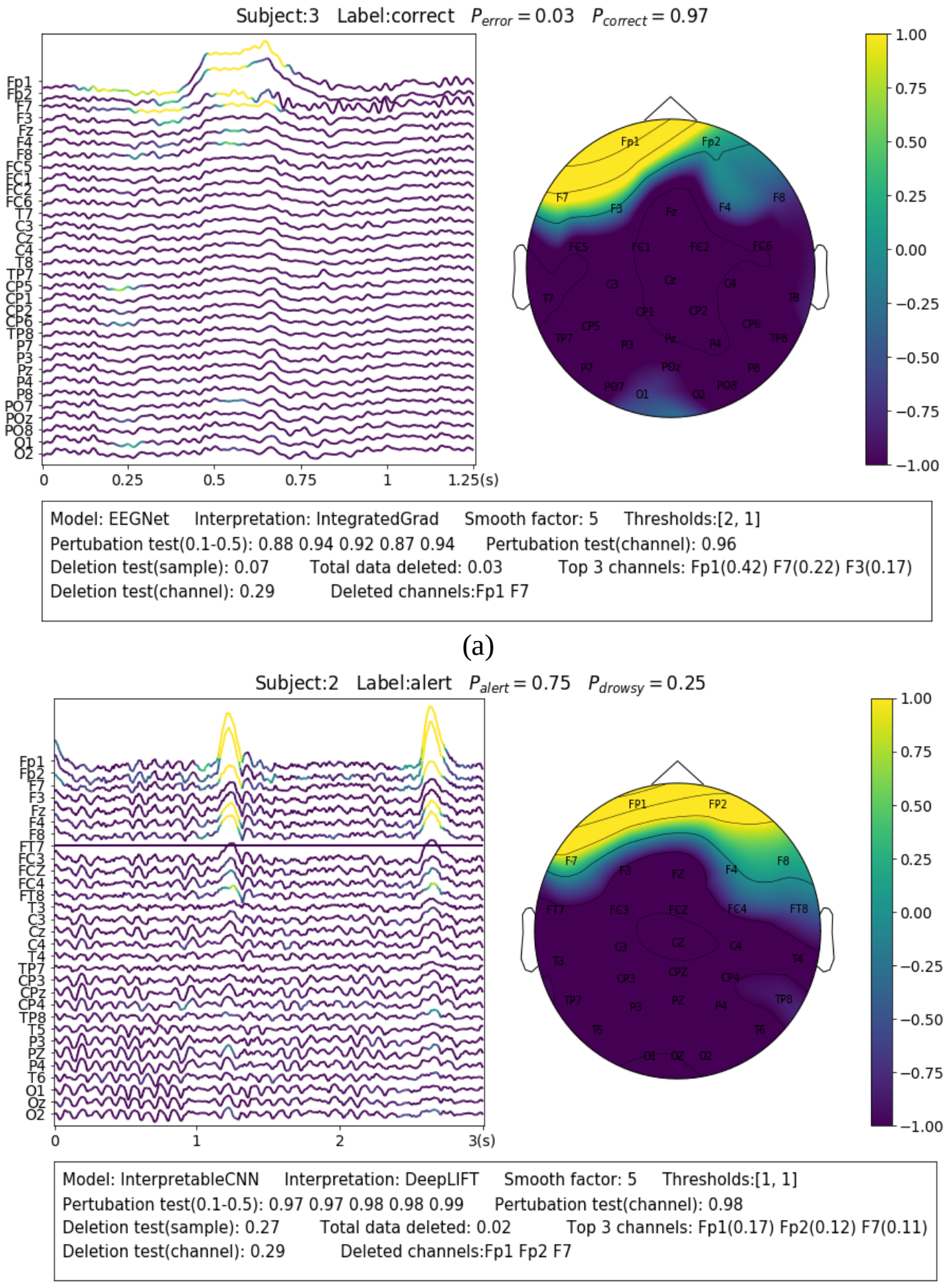

Fig. 6. Visualization of the interpretation results for two samples containing eye blink artifacts

## VII. Discussion

In this paper, we investigated the topic of applying deep learning interpretability to EEG signal recognition. Despite the wide application of deep learning, there are yet no guidelines or recommendations on how to interpret the results, what methods should be used and how accurate they can reflect the model decisions for EEG-based BCI. In order to fill this research gap, we conducted quantitative studies to evaluate existing interpretation techniques and explored the best practice of interpreting deep learning designed for EEG signal recognition.

The results revealed the importance of selecting a proper interpretation technique in the first step. Existing interpretation techniques, e.g., the saliency map method, despite being widely used in existing work for interpreting learned EEG patterns [33], could actually fail to outperform baseline with randomly generated contribution maps in certain circumstances. In addition, we also find that the quality of the interpretation results is inconsistent for individual samples despite when a method with an overall good performance is used. Many factors, including model structure and types of features in the samples, could potentially affect the quality of the interpretation results. It is therefore recommended to conduct sample-wise evaluation to validate the results. By far as we know these findings not yet raised wide awareness in work interpreting deep learning for EEG-based BCI, e.g., [7, 24].

In order to make the interpretation results understandable, we proposed a few processing steps that can effectively enhance the visualization. Furthermore, we extensively used deep learning interpretability to explore how different types of noise and artifacts in the datasets can affect the model decisions. We show the benefits with example. The method allows us to come to the conclusion that different types of noise and artifacts should be treated differently based on a comprehensive understanding of the overall pattern learned from the dataset. By far as we know, this has not yet been realized in existing studies. The noise and artifacts defined in one scenario could become important features in another scenario. Deep learning interpretability could be potentially used as a powerful tool to discover the patterns underlying a dataset, so that a proper strategy can be specifically designed in the pre-processing pipeline.

## VIII. Conclusion

In this paper, we explored the best practice of applying deep learning interpretability to EEG-based BCI. Firstly, we surveyed existing deep learning interpretation techniques and shortlisted seven of them that can be applied to deep learning models with different structures. We designed evaluation metrics and tested them with two benchmark deep learning models on three different EEG datasets. The results revealed the importance of selecting a proper interpretation technique as the initial step. Secondly, we proposed a series of processing steps that allow the sample-wise interpretation results to be presented in an understandable and trusted way. Lastly, we used examples to illustrate how deep learning interpretability can benefit EEG-based BCI. Our work illustrates a promising direction of using deep learning interpretability to discover meaningful patterns from complex EEG signals.